\documentclass[10pt,twocolumn,letterpaper]{article}

\usepackage[pagenumbers]{iccv} 
\PassOptionsToPackage{table}{xcolor}
\usepackage{bm}
\usepackage{multirow}
\usepackage{multicol}
\usepackage{graphicx}
\usepackage{amsmath}
\usepackage{booktabs}
\usepackage{pifont}
\usepackage{tcolorbox}
\usepackage{color}
\usepackage{colortbl}

\definecolor{iccvblue}{rgb}{0.21,0.49,0.74}
\usepackage[pagebackref,breaklinks,colorlinks,allcolors=iccvblue]{hyperref}

\def\paperID{13050} 
\def\confName{ICCV}
\def\confYear{2025}

\title{EagleVision: Object-level Attribute Multimodal LLM for Remote Sensing}
\author{
Hongxiang Jiang\textsuperscript{1}, \,
Jihao Yin\textsuperscript{1,*}, \,
Qixiong Wang\textsuperscript{1}, \,
Jiaqi Feng\textsuperscript{1}, \,
Guo Chen\textsuperscript{1} \\
\textsuperscript{1}Beihang University\\
{\tt\small \{jianghongxiang, jihaoyin, fengjiaqi, chenguo777\}@buaa.edu.cn, wangqixiong@xiaohongshu.com}
}

\begin{document}
\twocolumn[{
\maketitle
\vspace{-2em}
\begin{center}
    \centering
    \includegraphics[width=0.98\textwidth]{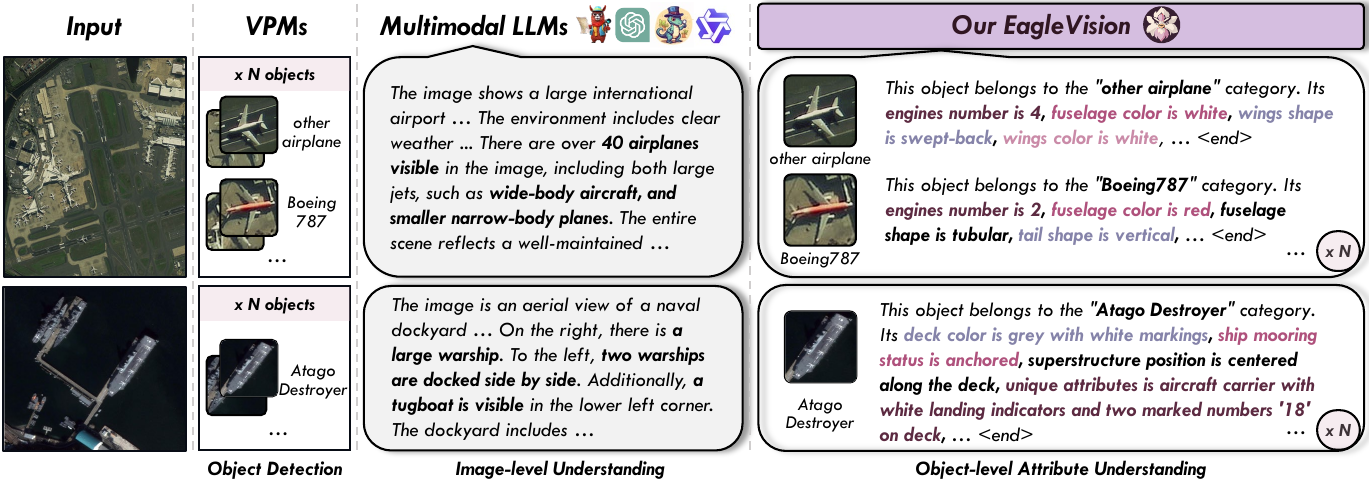}
    \vspace{-2mm}
    \captionof{figure}{\textbf{EagleVision for object-level attribute understanding.} In contrast to visual perception models (VPMs) and MLLMs, which contribute little to object-level comprehension in remote sensing, EagleVision outperforms in object attribute understanding, covering various attributes of all detected objects. The prompt for generating the MLLMs results is shown in the Appendix.}
    \label{intro}
\end{center}
}]

\begin{abstract}
Recent advances in multimodal large language models (MLLMs) have demonstrated impressive results in various visual tasks. However, in remote sensing (RS), high resolution and small proportion of objects pose challenges to existing MLLMs, which struggle with object-centric tasks, particularly in precise localization and fine-grained attribute description for each object. These RS MLLMs have not yet surpassed classical visual perception models, as they only provide coarse image understanding, leading to limited gains in real-world scenarios. To address this gap, we establish \textbf{EagleVision}, an MLLM tailored for remote sensing that excels in object detection and attribute comprehension. Equipped with the Attribute Disentangle module, EagleVision learns disentanglement vision tokens to express distinct attributes. To support object-level visual-language alignment, we construct \textbf{EVAttrs-95K}, the first large-scale object attribute understanding dataset in RS for instruction tuning, along with a novel evaluation benchmark, \textbf{EVBench}. EagleVision achieves state-of-the-art performance on both fine-grained object detection and object attribute understanding tasks, highlighting the mutual promotion between detection and understanding capabilities in MLLMs. The code, model, data, and demo will be available at \href{https://github.com/XiangTodayEatsWhat/EagleVision}{https://github.com/XiangTodayEatsWhat/EagleVision}.
\end{abstract}

\section{Introduction}
\label{sec:intro}
In recent years, the emergence of large language models (LLMs) \cite{DBLP:conf/nips/Ouyang0JAWMZASR22,touvron2023llama} has significantly impacted the research community, demonstrating remarkable achievements in following human instructions. With the integration of multimodal input (e.g., images, videos, or tables) and instruction tuning data, these models have further gained impressive visual reasoning skills, often referred to as multimodal large language models (MLLMs) \cite{DBLP:conf/nips/LiuLWL23a,DBLP:conf/cvpr/LiuLLL24,achiam2023gpt,chen2024internvl}. These methods have established strong alignments between vision-language modalities for performing a wide range of visual tasks, such as visual understanding and grounding \cite{DBLP:conf/eccv/MaJWYQ24}.

Although these general-purpose MLLMs have already been widely applied in various vertical domains, they are still in the early stages of remote sensing (RS). Some research like RSGPT \cite{hu2023rsgpt} and GeoChat \cite{DBLP:conf/cvpr/KuckrejaDNDKK24} only explored multitask conversations, performing tasks including scene classification and image description similar to natural images. In fact, the object-level interpretation is more practical in RS domain, but existing MLLMs struggle with precise object detection and object understanding. Especially for more profound, fine-grained object-centric tasks, both MLLMs and traditional object detection methods demonstrate severe limitations. Specifically, as illustrated in Fig.\,\ref{intro}, classical visual perception model (VPMs) for object detection can only locate objects relying on predefined categories \cite{DBLP:journals/pami/RenHG017,DBLP:conf/eccv/CarionMSUKZ20,zhou2019objects}. However, this approach proves insufficient in practical applications due to the lack of interpretability, especially when object types are unknown or novel, often resulting in vague labels such as “other-airplane” without other meaningful comprehension. Similarly, suffering from high resolution of RS images and small proportion of objects, MLLMs generally provide sparse and coarse captions such as “over 40 airplanes visible” or “smaller narrow-body planes”. These models struggle to describe fine-grained attributes for each object, leading to deficient object-centric understanding and no effective improvements in localization. Therefore, enhancing object-level attribute comprehension is a critical step in advancing RS MLLMs, contributing to the expansion of applications.

Motivated by this, we present EagleVision, a novel object-level attribute multimodal large language model for RS, capable of both object localization and fine-grained property description. To ensure object-level attribute comprehension in EagleVision, we propose an Attribute Disentangle module to obtain disentangled object vision tokens via orthogonal subspace learning. In contrast to original tokens, which inherently mix multiple attributes and tend to express global content, these disentangled tokens could explicitly capture distinct attribute features, thus facilitating further understanding fine-grained properties of objects. To support EagleVision training on object attribute understanding task, we build the EVAttrs-95K dataset for instruction tuning, aligning object visual features with their corresponding detailed descriptions. Specifically, we design an innovative annotation pipeline, providing open-ended annotations of detailed attributes for 95.1k objects across the FAIR1M, MAR20, and ShipRSImageNet datasets. Finally, we present EVBench, the first benchmark for evaluating object attribute understanding capability in RS. Experimental results demonstrate that EagleVision not only achieves state-of-the-art performance on object detection, improving mAP by 11.2\%, 2.7\% and 0.3\% on three datasets respectively, but also shows significant advantages in EVBench. The key contributions are as follows.

\begin{itemize}
    \item We build and fine-tune a novel MLLM architecture, \textbf{EagleVision}, which incorporates both object detection and object attribute understanding.
    \item To overcome attribute mixture of object vision tokens to ensure fine-grained comprehension in EagleVision, we present \textbf{Attribute Disentangle}, which facilitates disentangled attribute learning, benefiting multiple tasks.
    \item For the first time, we develop a large-scale remote sensing object attribute understanding dataset, \textbf{EVAttrs-95K} and an evaluation benchmark, \textbf{EVBench}, which provides comprehensive support and demonstrates the outstanding performance of our EagleVision.
\end{itemize}

\section{Related Work}
\label{sec:relatedwork}

\subsection{Visual Perception Models}
\begin{table}
\renewcommand{\arraystretch}{1.0}
    \setlength{\tabcolsep}{3pt}
    \centering
    {\small
    \begin{tabular}{lcc}
        \toprule
        Method &Reference &Description\\
        \midrule
        \rowcolor{gray!20}  
        \multicolumn{3}{l}{\textit{Visual Perception Models}} \\  
        \midrule
        DETR   (\textit{ECCV 2020})  & \ding{55}      & \ding{55}  \\
        KFIoU (\textit{ICLR 2023})  & \ding{55}      & \ding{55} \\
        PolyFormer (\textit{CVPR 2023})  & \ding{51}      & \ding{55} \\
        Grounding-DINO (\textit{ECCV 2024})    & \ding{51}      & \ding{55} \\
        \midrule
        \rowcolor{gray!20}  
        \multicolumn{3}{l}{\textit{Multimodal LLMs}} \\  
        \midrule
        LLaVA  (\textit{NIPS 2023})   & \ding{55}     & Image \\
        LLaVA-Grounding (\textit{ECCV 2024})  & \ding{51}     & Sparse \\
        GeoChat (\textit{CVPR 2024})  & \ding{51}     & Sparse \\
        GPT-4o                        & \ding{51}     & Sparse \\
        \midrule
        \rowcolor{purple!5}  
        \textbf{EagleVision}             & \ding{55}       & Dense \\
        \bottomrule
    \end{tabular}
    }
    \caption{\textbf{Comparison of related works.} “Reference” indicates whether the reference text is required for detection or segmentation. “Image” and “Sparse” refer to perform image-level and sparse object-level understanding.}
    \vspace{-5mm}
    \label{related works}
\end{table}

Traditional visual perception models, such as object detection, primarily focus on object localization and classification. Representative work series include anchor-based Faster R-CNN \cite{DBLP:journals/pami/RenHG017,DBLP:journals/pami/HeGDG20,DBLP:conf/cvpr/CaiV18}, anchor-free CenterNet \cite{zhou2019objects}, and transformer-based DETR \cite{DBLP:conf/eccv/CarionMSUKZ20,DBLP:conf/iclr/ZhuSLLWD21,DBLP:conf/iclr/0097LL000NS23}. Based on these works, remote sensing detection methods further emphasize addressing challenges like arbitrary rotations and small object sizes, exemplified by two-stage Oriented R-CNN \cite{DBLP:conf/iccv/Xie0WYH21} and one-stage R3Det \cite{DBLP:conf/aaai/YangYFH21}. Despite improved detection performance, they share a common issue: only supporting coarse category predictions and lacking fine-grained understanding of each object. In other words, an object can only be classified as an "airplane", without any detailed interpretation. In practical remote sensing applications, this issue makes it challenging to discover and analyze new types of objects, also hampering the improvement of detection capability. 

\begin{figure*}[t]
   {\centering
   \includegraphics[width=0.97\textwidth]{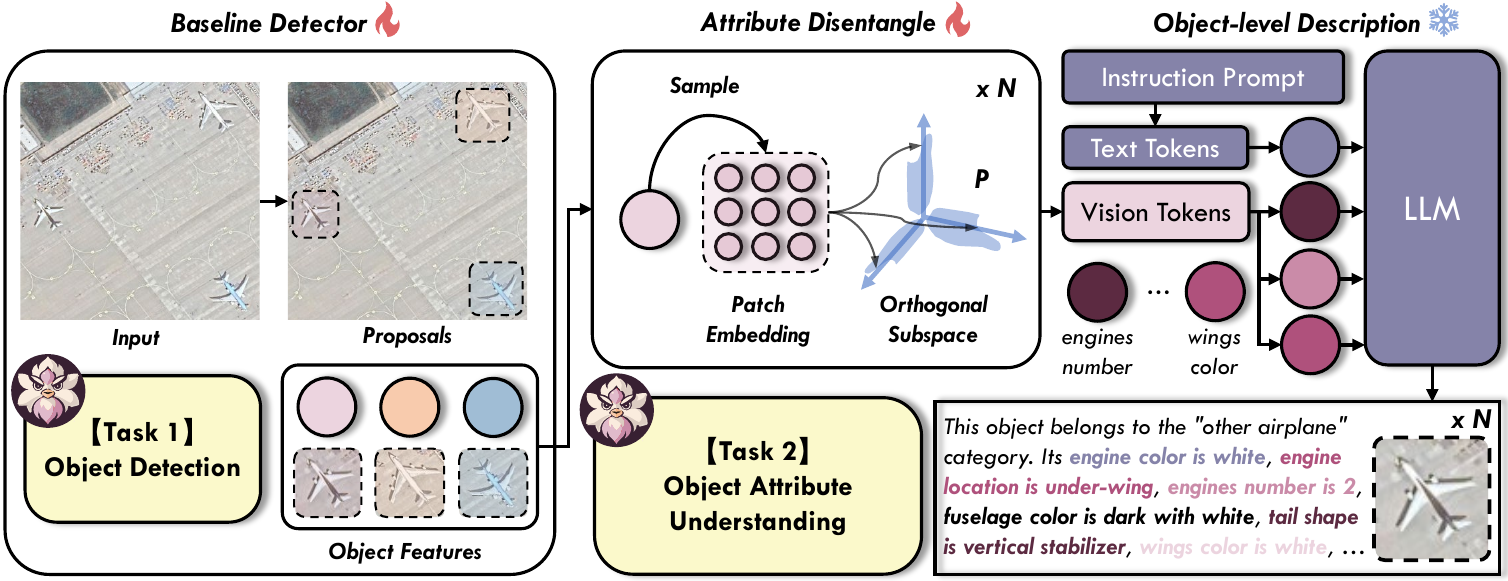}
   \caption{
      \textbf{The overall architecture of EagleVision.} EagleVision consists of three main components: Baseline Detector, Attribute Disentangle and Object-level Description, enabling object detection and object attribute understanding tasks.
   \label{framework}
   \vspace{-5mm}
   }}
\end{figure*}

Recently, VPMs for visual grounding (VG) and referring expression segmentation (RES), such as PolyFormer \cite{DBLP:conf/cvpr/LiuDCZSMM23} and Grounding-DINO \cite{DBLP:conf/eccv/LiuZRLZYJLYSZZ24}, have attracted significant attention due to their flexibility beyond predefined categories. However, these methods heavily rely on the reference text of known objects (e.g., "the blue airplane on the right side") for matching and localization. In fact, such approaches are more suited for interactions with specific objects, but not for remote sensing scenarios, as they fail to understand a wide range of unknown land-cover objects and cannot enhance detection of them.

It is worth noting that OvarNet \cite{DBLP:conf/cvpr/ChenJHTGCX23}, TAP \cite{DBLP:conf/eccv/PhamKLDCTS22}, etc. also attempt to identify the attributes of detected object in natural images. Nevertheless, these models involve a relatively complex multi-stage training process, and completely depend on the contrastive retrieval of CLIP \cite{DBLP:conf/icml/RadfordKHRGASAM21} without free-form descriptions, which struggle to generalize effectively to the remote sensing domain.

To address these limitations, EagleVision aims to perform reference-agnostic detection and fine-grained attribute understanding for each object. This dense understanding of objects also significantly improves detection capability.

\subsection{Multimodal Large Language Models} 

Empowered by the recent advancements in LLMs, MLLMs have demonstrated exceptional capabilities in visual understanding. LLaVA \cite{DBLP:conf/nips/LiuLWL23a}, an early MLLM, introduces a novel learning paradigm and instruction-tuning data construction, widely adopted and extended in subsequent works \cite{DBLP:conf/cvpr/LiuLLL24,DBLP:conf/eccv/ChenLDZHWZL24}. However, these related research primarily focuses on global image understanding, 
demonstrating limited capability in local object comprehension and visual perception. With the introduction of visual grounding modules, models like LLaVA-Grounding \cite{DBLP:conf/eccv/ZhangLLRZLHGLLY24} achieve object localization and understanding based on reference information. Unfortunately, MLLMs cannot produce satisfactory detection, with low recall rates \cite{jiang2024chatrex}. This deficiency causes a sparse object understanding, manifested in two aspects: sparse number, the missing understanding of many critical objects, and sparse attributes, leading to rough description. Even state-of-the-art models face similar challenges, such as open-source Qwen2-VL \cite{wang2024qwen2} and closed-source Gemini \cite{team2023gemini,team2024gemini}.

In the field of remote sensing, existing MLLMs, following general-domain architectures, are particularly difficult to solve the problem of sparse object comprehension due to the high image resolution and small proportion of objects. For instance, RSGPT \cite{hu2023rsgpt} concentrates on image-level QA tasks, GeoChat \cite{DBLP:conf/cvpr/KuckrejaDNDKK24} aims to build a versatile remote sensing MLLM with richer vertical domain data, and RSUniVLM \cite{liu2024rsunivlm} is proposed to extend RES and multi-image dialogue tasks. All of these methods could not address the inherent limitations in object-level understanding and detection.

Therefore, EagleVision, as the first object-level attribute MLLM in remote sensing, is proposed for dense understanding of over 60 fine-grained attributes for each object, while also ensuring precise detection capabilities. In summary, a comparison between EagleVision and the above related works is shown in Table. \ref{related works}.

\section{Method}
\label{method}

As illustrated in Fig. \ref{framework}, EagleVision firstly extracts object features and perform object detection via Baseline Detector. Then, to enable the original entangled features to express different properties, Attribute Disentangle module is introduced to produce attribute-separated vision tokens through orthogonal subspace learning. Finally, leveraging the LLM, Object-level Description accomplishes object attribute understanding. During training on our EVAttrs-95K, all the losses are calculated to update the whole vision part. 

\subsection{Baseline Detector}

For the input image $\bm{X_v}\in\mathbb{R}^{H\times W \times 3}$, we use the baseline detector to extract ROI features $\bm{F_v}=f(\bm{X_v}; \theta)$, where $\bm{F_v}\in\mathbb{R}^{N\times H'\times W' \times C}$, $N$ is the number of proposals, $H'$ and $W'$ are height and width of the ROI feature map, $f$ denotes any single-stage or two-stage detector, and $\theta$ represents the corresponding parameters. To achieve object detection, we retain the relevant modules $f_{cls}$ and $f_{reg}$ for classification and bounding box regression respectively. After feeding $\bm{F_v}$ into them, the final detection results could be obtained. According to these results, we further select the ROI features of $N_{pos}$ foreground objects as the object features $\bm{F_v^{pos}}\in\mathbb{R}^{N_{pos}\times H'\times W' \times C}$ for subsequent process.

For optimization, we calculate detection losses $\mathcal{L}_{d}$ following classical detector, including cross-entropy loss, L1 loss or RotatedIoU loss \cite{DBLP:journals/tmm/MaSYWWZX18}. All parameters in the detector are trainable. 

\subsection{Attribute Disentangle}
\label{module:ad}
Instead of directly inputting $\bm{F_v^{pos}}$ as the vision token to the LLM, to provide more sufficient object information, we firstly sample the neighborhood features of the object to obtain the patch embedding $\bm{E_v}\in\mathbb{R}^{N_{pos}\times (2s+1)\times(2s+1) \times C}$, where $s\in\mathbb{N}$. Specifically, for two-stage detector, the ROI feature size could be adjusted to $2s+1$, and then $\bm{E_v}$ is output. In comparison, $\bm{F_v^{pos}}$ from single-stage detector is defined with $H'=W'=1$, which adopts the center feature. Thus, for these objects, we determine their centers $R$ and sample the neighborhood around each center as follows:
\begin{equation}
\label{eq:patch embedding}
\begin{aligned}
    R&=\{r_i\}_{i=1,2,...,N_{pos}}, \:\:\: r_i=(x_i, y_i)\\
    S_i&=\{(x_i+s_x, y_i+s_y)|s_x, s_y\in\left[-s,s\right]\},
\end{aligned}
\end{equation}
where $S_i$ represents the neighborhood set of center $r_i$, for selecting corresponding features as $\bm{E_v}$.

Since the extracted $\bm{E_v}$ mixes various attribute features, lacking the ability to represent details, it tends to prompt the LLM to generate a global object caption, rather than specific properties. Therefore, to enable vision tokens to explicitly express different attributes for fine-grained understanding, we further introduce disentanglement learning, inspired by \cite{bengio2013representation,DBLP:conf/eccv/SarhanNEA20,DBLP:conf/icml/ChaT23}. We adopt the orthogonal subspace learning to disentangle the features across each property. In details, a set of orthogonal basis, $p_1$, $p_2$, ...,  $p_n$, is learned to span a hyperplane $\mathcal{P}=span\{p_1, p_2, ...,  p_n\}$, referred to as orthogonal subspace, where each basis represents a distinct attribute space. $n$ is the number of basis. The patch embedding $\bm{E_v}$ are then projected onto these basis to obtain disentangled features $\bm{T_v}\in\mathbb{R}^{N_{pos}\times n \times C}$, which is the final vision tokens, consisting of $n$ independent tokens. The specific implementation is depicted as following:

\begin{equation}
\label{eq:osr}
\begin{aligned}
    \bm{T_v}&=cat(\bm{T_{v}^{1}},\bm{T_{v}^{2}},...,\bm{T_{v}^{n}})\\
\bm{T_{v}^{k}}&=c_kp_k,\;c_k=\sum_{i}^{2s+1}\sum_{j}^{2s+1}\bm{E_{v}^{i,j}}p_{k}^{T},
\end{aligned}
\end{equation}
where $p_{k}\in\mathbb{R}^{1\times C}$ is the learned parameters, $\bm{E_{v}^{i,j}}\in\mathbb{R}^{N_{pos}\times C}$, $i$ and $j$ represent the indexes of the patch embedding. $cat$ represents the concatenation of tensors.

In the above process, for constraining the learnable parameter $p$ to ensure the orthogonality of the basis, namely $p_ip_{j}^{T}=0$ when $i\ne j$, we introduce the following orthogonality losses $\mathcal{L}_{o}$:
\begin{equation}
\label{eq:Lo}
\begin{aligned}
\mathcal{L}_{o}&=\frac{2}{n\times(n-1)}\sum_{i=1}^{n}\sum_{j>i}^{n}|p_ip_{j}^{T}|.
\end{aligned}
\end{equation}
To guide the disentangled representation $c_k$ in the attribute space obtained by Eq.\ \ref{eq:osr} to correctly express the corresponding attributes, the mutual information $I$ is maximized between $c_k$ and the attribute token $\bm{T_a^k}$ encoded from the groundtruth attribute:
\begin{equation}
\label{eq:info}
\begin{aligned}
\mathcal{L}_{a}&=-\frac{1}{n}\sum_{k}^{n}I(c_k, \bm{T_a^k}).
\end{aligned}
\end{equation}
This goal, based on information theory, is prevalent in various representation learning or correlation constraints. Here, it specifically emphasizes one-to-one correspondence between vision tokens and their associated attributes. Due to the intractability of Eq. \ref{eq:info}, encouraged by \cite{DBLP:conf/nips/ChenCDHSSA16}, we optimize its variational lower bound:
\begin{equation}
\label{eq:lowerbound}
\begin{aligned}
\mathcal{L}_{a}&=\frac{1}{n}\sum_{k}^{n}(q(\bm{T_a^k};\varphi) - c_k)^2\\
&=-\frac{1}{n}\sum_{k}^{n}\mathbb{E}_{\bm{T_a^k}}[\mathbb{E}_{c_k\sim P(c_k|\bm{T_a^k})}[log(Q(c_k|\bm{T_a^k})]]\\
&\geq-\frac{1}{n}\sum_{k}^{n}I(c_k, \bm{T_a^k}) + H(c),
\end{aligned}
\end{equation}
where $Q(c_k|\bm{T_a^k})\sim\mathcal{N}(q(\bm{T_a^k};\varphi),I)$ is the variational distribution. It should be noted that $\bm{T_a^k}$ is only used for training in Eq. \ref{eq:lowerbound} and does not exist during testing. Ultimately, thanks to the proposed attribute disentangle module incorporating the decorrelating transformation, the independence between features is enhanced which could express different attributes, contributing to visual-language alignment. This will be further confirmed in Sec. \ref{ablation text}.

\subsection{Object-level Description}
Eventually, text tokens $\bm{T_q}$ encoded from the instruction prompt and vision tokens $\bm{T_v}$ are concatenated and fed into the frozen LLM to generate object-level descriptions. It realizes the dense attribute understanding task for each object, which is formulated as $\bm{Y}=g(\bm{T_v}, \bm{T_q};\phi)$, where $\bm{Y}$ is the response of LLM $g$ and $\phi$ denote the frozen parameters. Based on these response $\bm{Y}$ and groundtruth attribute descriptions $\bm{\hat{Y}}$ in the EVAttrs-95K dataset, we calculate the language loss $\mathcal{L}_{q}$ with the simple next-token prediction loss. Only the visual component of EagleVision is optimized. In fact, this step aligns the object-level visual features with the LLM word encoding, ensuring that the vision part in EagleVision is compatible with the frozen LLM, similar to the pre-training in LLaVA. In addition, while focusing on the attribute understanding task, $L_q$ indirectly improves visual feature extraction, which is more consistent with the object characteristics and beneficial to detection.

As a result, our EagleVision achieves more accurate detection performance than baseline detector, and facilitates mutual enhancement between detection and object attribute understanding. The complete loss function is as follows, each $\lambda$ represents the weight coefficient for a specific loss:
\begin{equation}
\label{eq:overall}
\begin{aligned}
\mathcal{L}_{overall}&=\lambda_d\mathcal{L}_{d} + \lambda_o\mathcal{L}_{o} + \lambda_a\mathcal{L}_{a} + \lambda_q\mathcal{L}_{q}.
\end{aligned}
\end{equation}

\begin{table}
\renewcommand{\arraystretch}{1.0}
    \setlength{\tabcolsep}{6pt}
    \centering
{\small
    \begin{tabular}{lccc}
        \toprule
        Data  & FAIR1M & MAR20 & ShipRSImageNet\\
        \midrule
        Size         & 59.8k  & 22.3k  & 13.0k      \\
        Train        & 44.2k  & 7.8k  & 10.1k      \\
        Test         & 15.6k  & 14.5k  & 2.9k     \\
        Attr Num & $\sim$25 & $\sim$24  & $\sim$28   \\
        \bottomrule
    \end{tabular}
    }
    \caption{\textbf{Source and distribution of the EVAttrs-95K dataset.} $\sim$ indicates approximation of the average number of attributes.}
    \vspace{-4mm}
    \label{evattrs}
\end{table}

\subsection{EVAttrs-95K Generation Pipeline}
To equip EagleVision with robust object detection and attribute understanding capability, we construst EVAttrs-95K dataset with detailed attributes of 95.1k objects. The annotation process diagram, predefined attributes, annotation example, and detailed prompt design are provided in the Appendix. The following is the complete process.

\textbf{Dataset Preprocess.} Considering that object attributes could better promote fine-grained object detection task, we first select images from the train and validation set of FAIR1M-v1.0 \cite{sun_fair1m_2022} and ShipRSImageNet \cite{DBLP:journals/staeors/ZhangZWFH21}, and the train and test set of MAR20 \cite{wenqi2024mar20}. Among them, FAIR1M contains five major categories: airplane, ship, vehicle, court, and road, which are subdivided into 37 subcategories, ShipRSImageNet contains 50 types of ships, and MAR20 contains 20 types of airplanes. Moreover, we crop all the patches of airplane and ship objects from these images, and predefine 24 and 38 attribute names for airplanes and ships.

\textbf{Two-stage Annotation.} Given the patches and predefined attribute names, we employ a two-stage data engine. In the first stage, Qwen2-VL-72B is used to annotate all samples, followed by GPT-4o in the second stage to annotate low-quality samples, typically caused by small or blurred objects. The same prompt is applied in both stages, and the output is restricted to the formatted JSON. Specifically, we add an additional confidence, which needs to be given by MLLMs as the certainty of its annotation, ranging from 0 to 1. In the second stage, we re-annotate the samples with confidence less than 0.5, generating confidence again for subsequent human refinement. In the first stage, we deploy Qwen2-VL-72B locally using 4 Nvidia A100 GPUs, with a total annotation time of approximately 316 hours. In the second stage, the annotation time is around 8 hours.

\textbf{Human Refinement.} Although the automated process successfully annotate most object attributes, some results remain ambiguous. Therefore, we meticulously review all the annotations with confidence below 0.7, correcting the attribute descriptions that are obviously inconsistent with the image, and removing uncertain annotations such as “unable to annotate without clear visual information”. A brief distribution of EVAttrs-95K is shown in Table. \ref{evattrs}.

\subsection{EVBench}
\label{evbench}
To evaluate model performance on object-level attribute understanding task efficiently, we propose EVBench, which employs a meticulously curated evaluation strategy for attribute descriptions generated by MLLMs. It encourages accurate and comprehensive prediction of each attribute of every object in an image, and provides an effective assessment that highlights the performance gaps between MLLMs.

\textbf{Data Splits.} Firstly, we clarify the data splits of our EVAttrs-95K in Table.\ \ref{evattrs}. For FAIR1M, we manually divide the train and test set in a ratio of 3:1 from original trainval set. For MAR20, we inherit its train and test set. Since the test set of ShipRSImageNet is not publicly available, we use the original train and validation set for training and testing.

\begin{table*}
    \centering
    \renewcommand{\arraystretch}{0.9}
    \setlength{\tabcolsep}{8pt}
    {\small
    \begin{tabular}{lccccc}
        \toprule
        
        Method          & Patch Embedding & Vision Token & LLM        & mAP     &Score \\
        \midrule
        \multirow{4}{*}{EagleVision-1B$\dag$}  & $1\times1$ & \multirow{4}{*}{Entangled} &\multirow{4}{*}{Qwen2-0.5B-Instruct \cite{qwen2}} & 56.8 & 56.8 \\
        &    $3\times3$ & &   & 59.5             &63.9  \\
        &    $5\times5$ & &   & \textbf{64.4}    &\textbf{65.1}  \\
        &    $7\times7$ & &   & 62.2             &64.3  \\
        \midrule
        \multirow{3}{*}{EagleVision-1B$\dag$} & \multirow{3}{*}{$5\times5$} & Entangled & \multirow{3}{*}{Qwen2-0.5B-Instruct \cite{qwen2}}  &64.4 &65.1 \\
        &     & Disentangled             &   &\textbf{67.0}         &66.2  \\
        &     & Orthogonal Disentangled  &   &66.4 &\textbf{67.4}  \\
        \midrule
        EagleVision-1B$\dag$        & \multirow{5}{*}{$5\times5$}  & \multirow{5}{*}{Orthogonal Disentangled}   & Qwen2-0.5B-Instruct \cite{qwen2}  & 66.4 & 67.4  \\
        EagleVision-1B        &   &  & Qwen2-0.5B-Instruct \cite{qwen2}  &67.1  &69.3  \\
        EagleVision-2B        &   &    & InternLM2-1.8B \cite{cai2024internlm2}       & 71.6 & 68.6  \\
        EagleVision-4B        &   &    & Phi-3-Mini-128K-Instruct \cite{abdin2024phi}    & 73.3     & 69.5  \\
        EagleVision-7B        &   &    & InternLM2.5-7B-Chat \cite{cai2024internlm2}         & \textbf{74.6}     & \textbf{69.9}  \\
        \bottomrule
    \end{tabular}
    }
    \caption{\textbf{Results of ablation study.} The table shows the metrics of two benchmarks on object detection and attribute understanding, where “Score” is from the GPT-assisted evaluation in Sec. \ref{evbench}. $\dag$ indicates that RTMDet is as baseline detector, otherwise Oriented R-CNN.}
    \vspace{-4mm}
    \label{ablation}
\end{table*}

\textbf{Response Preprocessing.} Furthermore, we perform object attribute understanding for all images of the test set, and obtain the response of $N'$ objects $\{\bm{Y'_{i}}\}_{i=1,2,...,N'}$, which are empty for undetected objects. To rigorously evaluate the results of each attribute, we convert the non-empty $\bm{Y'_{i}}$ and corresponding groundtruth $\bm{\hat{Y}'_{i}}$ into JSON $\mathcal{D}_i$ and $\hat{\mathcal{D}}_i$, where key is the name and value is the description of attributes.

\textbf{Evaluation Strategy.} To assess the object completeness in MLLMs attribute comprehension, we first introduce the recall metric, which quantifies the proportion of objects with non-empty responses over the total number of objects. Recall serves as an indicator of MLLM's capability to detect objects effectively, ensuring accurate execution of object-level tasks. Then, we consider evaluating the accuracy of attribute understanding. Since object attribute understanding task requires open-ended answers generation, and the values of $\bm{Y'_{i}}$ and $\bm{\hat{Y}'_{i}}$ are uncertain, we employ a GPT-assisted evaluation strategy that could be consistent with human evaluation, which has been verified in most recent MLLM benchmarks \cite{DBLP:conf/nips/LiuLWL23a,DBLP:conf/nips/YinWCSLLH0S0SO23,DBLP:conf/eccv/LiuDZLZZYWHLCL24,DBLP:conf/icml/YuYLWL0WW24}. The model version selected as the evaluator is gpt-3.5-turbo-0125 and the prompt for evaluation is provided in the Appendix. According to the designed evaluation criteria, it compare the generated answer $\mathcal{D}_i$ and the reference answer $\hat{\mathcal{D}}_i$, the score of each attribute in each object could be obtained, ranging from 1 to 5. The final attribute score is the average score across all objects for a given attribute, scaled to a maximum score of 100 and the total score is the average of all attribute scores.

\section{Experiments}

\subsection{Implementation Details}

In our experiment, we report the results of the object detection and object attribute understanding on FAIR1M-v1.0, MAR20, and ShipRSImageNet datasets. For fair comparison, we adopt the same dataset processing for all methods unless otherwise specified. For FAIR1M-v1.0, we adopt single-scale training and testing strategy by cropping each image into 1024×1024 sub-images with a patch overlap of 200 pixels. For MAR20 and ShipRSImageNet, we directly rescale the origin images to 1024×1024 for experiments.

To be compatible with different requirements, our EagleVision includes models of four sizes: 1B, 2B, 4B and 7B. The LLMs are initialized with the corresponding language components from InternVL2 of 1B, 2B, 4B, and 8B.

\begin{table*}
    \centering
    \renewcommand{\arraystretch}{0.85}
    \setlength{\tabcolsep}{8pt}
    \small{
    \begin{tabular}{lcccccccc}
        \toprule
        \multirow{2}{*}{Method} & \multirow{2}{*}{ShipRSImageNet} & \multirow{2}{*}{MAR20} &\multicolumn{6}{c}{FAIR1M}  \\
        & & & Airplane & Ship & Vehicle & Court & Road & Mean \\
        \midrule
        \rowcolor{gray!20}  
        \multicolumn{9}{l}{\textit{One-stage Detector}} \\  
        \midrule
        RetinaNet \cite{DBLP:conf/iccv/LinGGHD17}    &20.1  &68.6   &37.7   &11.9 &10.8 &62.5 &21.0 &26.6   \\
        R3Det \cite{DBLP:conf/aaai/YangYFH21}      &23.8 &65.6    &39.0   &18.8   &18.2 &64.8 &30.8 &31.1 \\
        GWD \cite{DBLP:conf/icml/0005YMWZ021}        &26.7 &74.3    &40.2   &13.3   &13.2 &62.8 &26.1 &28.1 \\
        KLD \cite{DBLP:conf/nips/YangYYMWTY21}        &49.2 &80.8    &39.6   &13.2   &13.7 &63.8 &26.4 &28.3 \\
        FCOS \cite{DBLP:conf/iccv/TianSCH19}     &56.0  &80.2   &42.4   &23.8 &18.9 &66.9 &35.5 &34.1   \\
        S2ANet \cite{DBLP:journals/tgrs/HanDLX22}     &49.4  &42.6   &43.8   &23.0 &23.4 &65.7 &28.2 &34.7   \\
        TIOE-Det \cite{MING2023241}     &-  &-   &45.8   &16.9 &25.0 &69.9 &32.7 &35.2   \\
        RTMDet \cite{lyu2022rtmdet}     &59.2  &77.2   &44.5   &27.2 &\textbf{28.3} &\textbf{70.9} &34.3 &38.4   \\
        \midrule

        \rowcolor{gray!20}  
        \multicolumn{9}{l}{\textit{Two-stage Detector}} \\  
        \midrule
        Faster R-CNN \cite{DBLP:journals/pami/RenHG017}      &54.8 &75.0 &48.9    &21.4   &25.7 &65.5 &33.0 &36.8 \\
        Gliding Vertex \cite{DBLP:journals/pami/XuFWWCXB21}   &58.6 &80.3 &46.1    &21.4    &26.4 &67.3 &33.5 &36.5     \\
        ReDet \cite{DBLP:conf/cvpr/HanD0X21}               &53.9 &65.5  &47.2        &21.9    &25.3 &68.7 &30.4 &36.5    \\
        KFIoU \cite{DBLP:conf/iclr/00050ZYWY0023}               &37.5 &77.0  &44.4        &25.4  &19.2 &61.3 &26.8 &33.7    \\
        ROI Transformer \cite{DBLP:conf/cvpr/DingXLXL19}          &61.0 &82.5 &\textbf{50.8}    &24.1    &28.2 &68.3 &34.7 &39.2    \\
        Oriented R-CNN \cite{DBLP:conf/iccv/Xie0WYH21}      &63.4 &81.8 &46.0    &28.5    &26.0 &69.6 &35.8 &38.5     \\
        Oriented R-CNN* \cite{DBLP:conf/iccv/Xie0WYH21}     &- &- &53.6    &32.2    &38.9 &73.3 &38.2 &45.6   \\
        LSKNet* \cite{Li_2024_IJCV}             &- &- &53.6    &32.8    &\textcolor{purple!65}{\textbf{40.9}} &\textcolor{purple!65}{\textbf{76.6}} &40.8 &46.9    \\
        \midrule
        \rowcolor{purple!5}  
        \multicolumn{9}{l}{\textit{Ours}} \\  
        \midrule
        \textbf{EagleVision-1B}      &67.1 &82.7 &46.4    &28.6   &26.1 &69.7 &35.4 &38.6     \\

        \textbf{EagleVision-2B}      &71.6 &84.0 &50.3    &27.1    &26.6 &69.7 &31.7 &39.2     \\
        \textbf{EagleVision-4B}      &73.3 &84.3    &49.3    &29.0 &26.3 &68.0 &30.9 &39.0    \\
        \textbf{EagleVision-7B}      &\textbf{74.6} &\textbf{84.5}    &48.1    &\textbf{29.4} &27.6 &70.6 &\textbf{36.6} &\textbf{39.9}    \\
        \textbf{EagleVision-7B*}     &- &- &\textcolor{purple!65}{\textbf{54.4}}    &\textcolor{purple!65}{\textbf{33.3}}    &40.6 &76.5 &\textcolor{purple!65}{\textbf{41.2}} &\textcolor{purple!65}{\textbf{47.2}}    \\
        \bottomrule
    \end{tabular}
    }
    \caption{\textbf{Performance comparison on the object detection task.} * indicates the multi-scale training strategy by rescaling the images into three scales (0.5, 1.0, 1.5) and cropping each image into 1024×1024 with a patch overlap of 500 pixels.}
    \vspace{-5mm}
    \label{tab:od}
\end{table*}

All the models are implemented under MMRotate framework, and combined with the DeepSpeed \cite{DBLP:conf/sc/RajbhandariRRH20} engine to support LLM in our EagleVision. Following \cite{DBLP:conf/iccv/Xie0WYH21}, we train the models for 12 epochs on the FAIR1M dataset, and 36 epochs on the MAR20 and ShipRSImageNet datasets, with the AdamW \cite{DBLP:conf/iclr/LoshchilovH19} optimizer. We use 8 Nvidia A100 GPUs with a batch size of 8 for model training and testing. For more detailed configuration, see the Appendix.

\subsection{Ablation Study}
\label{ablation text}
In this section, we report the ablation study on the ShipRSImageNet to thoroughly investigate the effectiveness of the proposed method, as shown in Table. \ref{ablation}.

\textbf{Patch Embedding Size.} First, we directly input the original entangled vision tokens $\bm{E_v}$ from RTMDet into LLM, without applying Eq.\ \ref{eq:osr}, to explore the effect of the patch embedding size across four scale configurations. It can be seen that EagleVision-1B$\dag$ achieves a mAP of 56.8\% on object detection and a score of 56.8 on attribute understanding, only given the center token, namely $1\times1$. As the token size increases to $3\times3$, both mAP and score improve by 2.7\% and 7.1, respectively. With the $5\times5$ patch embedding, they further rise by 4.9\% and 1.2. When the patch size is increased to $7\times7$, suffering from irrelevant information around the object, the attribute understanding score decreases slightly by 0.8, which also affects detection with a 2.2\% reduction. Therefore, appropriately increasing the number of vision tokens could allow the LLM to receive more visual information, thereby improving both attribute comprehension and object detection, resulting in significant gains in both tasks. We finally choose a patch embedding size of $5\times5$.

\begin{figure}[t]
   {\centering
   \includegraphics[width=0.48\textwidth]{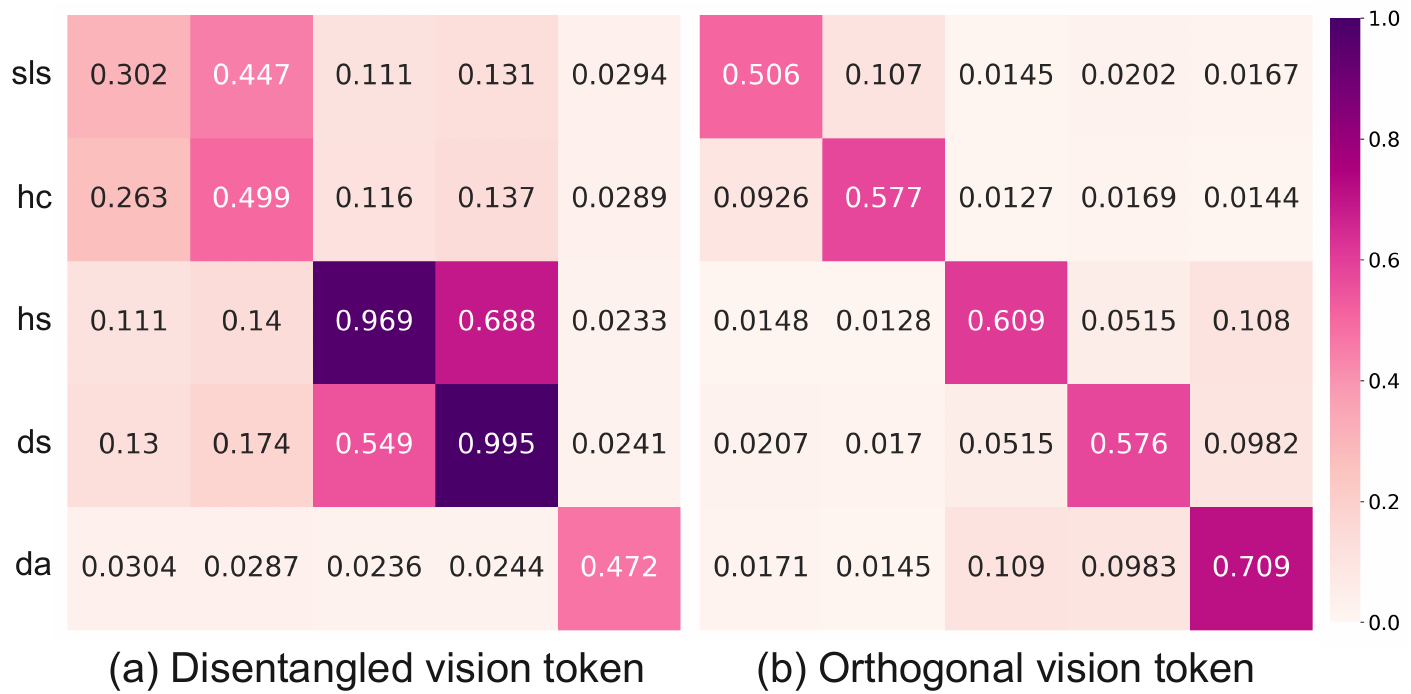}
   \caption{
      \textbf{Visualization of the correlation between vision tokens and attributes.} The horizontal axis represents different dimensions of vision tokens, and the vertical axis represents their attributes, where sls, hc, hs, ds, da denote ship-load-status, hull-color, hull-size, deck-structure, deck-accessories, respectively.
   }
   \vspace{-5mm}
   \label{correlation}
   }
\end{figure}

\textbf{Vision Token Type.} Then, we conduct validation on the performance of the disentangled vision tokens, introducing $\mathcal{L}_d$ presented in Eq.\ \ref{eq:lowerbound}. Leveraging the learned attribute-specific disentangled visual features, EagleVision achieves enhanced representational capacity, yielding notable improvements of 2.6\% and 1.1 in mAP and score, respectively. Furthermore, by incorporating orthogonal constraints $\mathcal{L}_o$ from Eq.\ \ref{eq:Lo}, the vision tokens exhibit superior disentangled properties, which facilitate more discriminative attribute understanding, improving 1.2 in score.

Intuitively, we visualize the disentanglement capability of our proposed vision tokens in Fig. \ref{correlation}, using measurement:
\vspace{-2mm}
\begin{equation}
\label{eq:corr}
\begin{aligned}
Corr(i,j)=\frac{min_{1 \leq i,j \leq n}|q(\bm{T_a^i};\varphi) - c_j|}{|q(\bm{T_a^i};\varphi) - c_j|},
\end{aligned}
\end{equation}
which comes from the variational lower bound we introduce in Sec. \ref{module:ad}, will be computed for all objects. $Corr(i,j)$ directly represents the correlation between the $i$-th dimension of the vision token and the $j$-th attribute, reflecting whether the token could express single attribute information. In addition, it converts the original absolute error $|q(\bm{T_a^i};\varphi) - c_j|$ into data ranging from 0 to 1, so that larger values indicate stronger correlation. As observed, although the disentangled vision tokens in (a) demonstrate partial independence, with increased performance, some attributes remain susceptible to confusion. For instance, the vision token in the fourth column shows a strong correlation of 0.995 with the deck-structure attribute, but also retains a high correlation of 0.688 with the hull-size. It is difficult to accurately understand both attributes based on such a mixed token. Benefiting from our orthogonal subspace learning, the orthogonal vision tokens in (b) achieve greater independence, enabling more precise understanding and further improving score. In comparison, the mAP only experiences a slight decline, so we still adopt the orthogonal disentangled vision tokens.

\begin{figure*}[t]
   {\centering
   \includegraphics[width=0.95\textwidth]{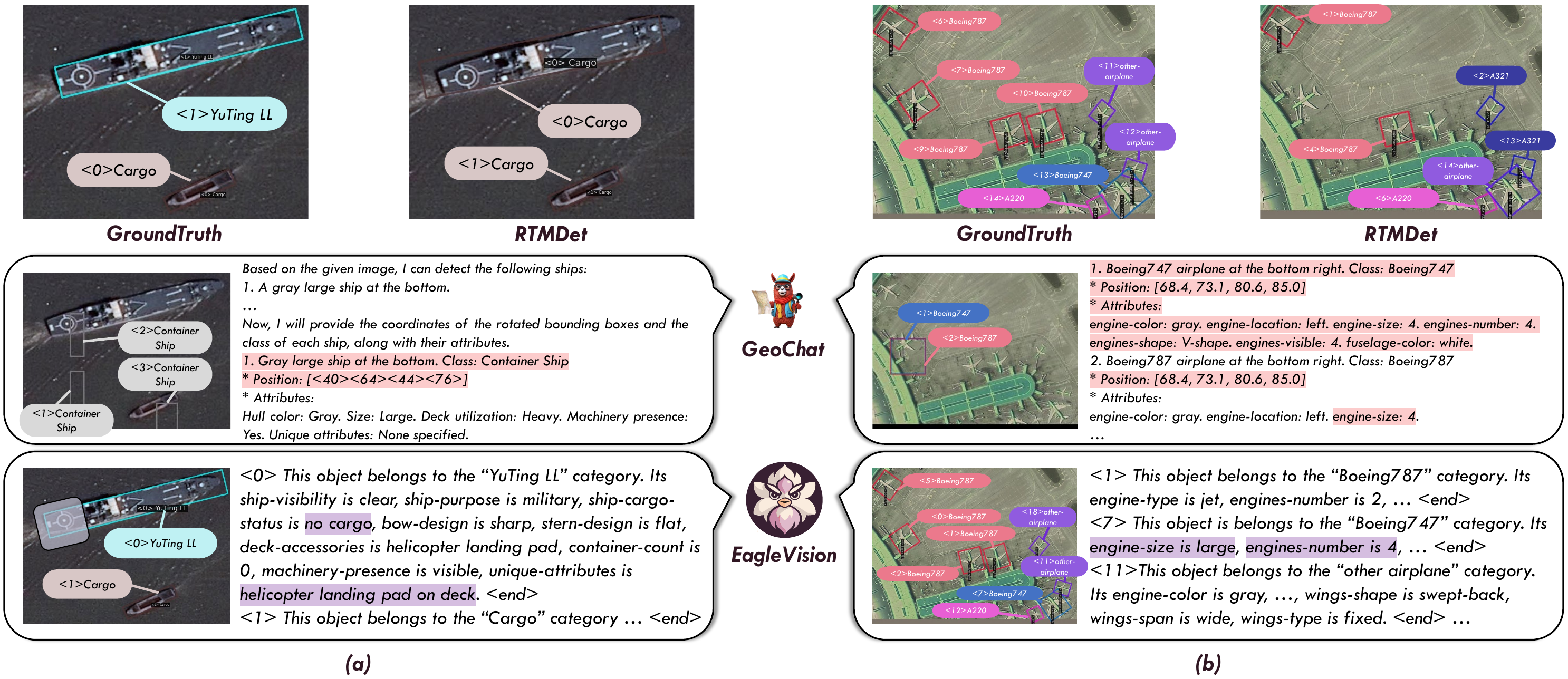}
   \vspace{-3mm}
   \caption{
     \textbf{Visualization results on ShipRSImageNet and FAIR1M datasets}. The results of RTMDet and GroundTruth only include detection, while the results of GeoChat are from the response to predefined prompt. In EagleVision, we highlight the crucial attribute understanding content, which promotes the correct detection of the object category.
   }
   \vspace{-4mm}
   \label{visualization}}
\end{figure*}

\textbf{Baseline Detector.} Exploring different types of baseline detectors, we replace the single-stage RTMDet with the two-stage Oriented R-CNN, yielding improvements of 0.7\% in mAP and 1.9 in score. This demonstrates the compatibility of EagleVision to various detectors. Given its superior performance, Oriented R-CNN is selected as the baseline detector for subsequent experiments.

\textbf{LLM Scaling.} Finally, we build four versions of EagleVision equipped with LLMs of different sizes. Thanks to the larger scale language module, the model's performance consistently improves, and it is particularly impressive that from 1B to 2B, the mAP is improved by 4.5\%, and from 2B to 4B, the mAP and score are both improved by 1.7\% and 0.9, and the best model, EagleVision-7B reaches a mAP of 74.6\% and score of 69.9.

\subsection{Task Evaluation}
To comprehensively illustrate the advantages of our EagleVision, we perform evaluation on multiple benchmarks of object detection and object attribute understanding tasks.

\textbf{Object Detection.} In terms of object detection task, we evaluated the performance of our EagleVision against 15 state-of-the-art detectors on three fine-grained object detection datasets. The results in Table. \ref{tab:od} show that EagleVision surpasses the baseline detector Oriented R-CNN across three datasets. Even the 1B version increases by 3.7\%, 0.9\%, and 0.1\% in mAP, respectively. Under the single-scale setting, our best EagleVision-7B outperforms all other methods with the mAP of 74.6\%, 84.5\% and 39.9\%. In particular, although we only annotate attributes of airplane and ship on FAIR1M, EagleVision not only yields gains of 4.3\% and 0.9\% in both, but also improve other categories by 1.6\%, 1.0\%, and 0.8\%. Under the multi-scale setting of FAIR1M, our method surpasses the state-of-the-art LSKNet by 0.3\%. Without additional computational overhead, EagleVision is compatible with any detector, maintains inference efficiency of detection, and brings stability improvements. This highlights the potential of MLLMs to enhance visual perception via object-level attribute understanding.

\begin{table}
    \centering
    \renewcommand{\arraystretch}{0.85}
    \setlength{\tabcolsep}{3pt}
    {\small
    \begin{tabular}{lcccccc}
        \toprule
        \multirow{2}{*}{Method} & \multicolumn{2}{c}{ShipRS} & \multicolumn{2}{c}{MAR20} &\multicolumn{2}{c}{FAIR1M}\\
        
         &Recall & Score & Recall & Score &Recall &Score \\
        \midrule
        \rowcolor{gray!20}  
        \multicolumn{7}{l}{\textit{General MLLMs}} \\  
        \midrule
        LLaVA-G \cite{DBLP:conf/eccv/ZhangLLRZLHGLLY24}  &0.5\%   &3.4   &1.8\%   &1.5    &1.2\%   &3.7     \\
        Qwen2-VL \cite{wang2024qwen2}          &8.2\%   &36.2  &52.5\% &42.2      &16.9\%     &40.3      \\
        InternVL2.5 \cite{chen2024expanding}           &9.7\%      &28.9  &21.8\% &44.3  &3.2\% &44.7        \\
        GPT-4o-mini \cite{achiam2023gpt}     &0.7\%     &38.0   &4.8\%  &45.7   &3.5\%    &39.9        \\
        
        \midrule
        \rowcolor{gray!20}  
        \multicolumn{7}{l}{\textit{Remote Sensing MLLMs}} \\  
        \midrule
        GeoChat \cite{DBLP:conf/cvpr/KuckrejaDNDKK24}            &1.6\% &22.1  &5.9\%  &19.8   &3.7\%   &23.5               \\
        LHRS-Bot \cite{DBLP:conf/eccv/MuhtarLGZX24}         &7.3\%   &37.8   &2.0\% &27.7  &2.5\%    &33.4          \\
        \midrule
        \rowcolor{purple!5}  
        \multicolumn{7}{l}{\textit{Ours}} \\  
        \midrule
        
        \textbf{EagleVision-1B} &77.3\%  &69.3   &91.6\%  &86.2  &\textbf{90.2\%} &75.0    \\
        \textbf{EagleVision-2B}  &77.1\% &68.8  &93.5\%   &88.8 &89.5\%  &76.2    \\
        \textbf{EagleVision-4B} &76.8\%  &69.5 &\textbf{94.3\%}   &88.4  &89.5\% &\textbf{76.3}  \\
        \textbf{EagleVision-7B}  &\textbf{79.0\%} &\textbf{69.9}    &92.8\% &\textbf{91.1}     &86.6\% &75.7  \\
        \bottomrule
    \end{tabular}
    }
    \caption{\textbf{Performance comparison on the object attribute understanding task.} “LLaVA-G” and “ShipRS” stands for LLaVA-Grounding and ShipRSImageNet.}
    \vspace{-5mm}
    \label{tab:oau}
\end{table}


\textbf{Object Attribute Understanding.} For attribute understanding task, we compare our EagleVision with 6 advanced MLLMs, as shown in Table \ref{tab:oau}. The results demonstrate the low recall of existing MLLMs in remote sensing scenarios, with significant object omission and hindering the acquisition of critical object attributes. Despite MAR20 containing more large-scale objects, only Qwen2-VL and InternVL2.5 achieve relatively better recall of 52.5\% and 21.8\%, while all models perform below 20\% on other datasets. In addition, these MLLMs exhibit consistently low scores, particularly on ShipRSImageNet, where GPT-4o achieves the highest score of only 38.0. Notably, remote sensing MLLMs, due to supervised fine-tuning (SFT) on limited domain tasks, manifest poorer generalization in object-level tasks. Only LHRS-Bot achieves a comparable performance on ShipRSImageNet with recall of 7.3\% and score of 37.8. In contrast, EagleVision demonstrates a substantial performance advantage. For example, EagleVision-7B achieves recall and score of 79.0\% and 69.9 on ShipRSImageNet, 92.8\% and 91.1 on MAR20, 86.6\% and 75.7 on FAIR1M, far outperforming other methods.

\subsection{Visualization}
Visualization examples are shown in Fig.\;\ref{visualization}. Compared to the inferior detection of the baseline detector RTMDet and the sparse object-level understanding and detection exhibited by the remote sensing MLLM GeoChat, EagleVision provides more accurate object detection and comprehensive object attribute description. It not only captures richer semantic information for unknown categories, such as “other-airplane”, but also enhances interpretability for correct detection through its identification of specific attributes. For instance, in Fig.\;\ref{visualization} (a), EagleVision obtains the attributes of “no cargo” and “helicopter landing pad on deck” of the “YuTing LL”, thereby clarifying that the object is not a cargo but rather a landing ship. In Fig. \ref{visualization} (b), all objects are correctly detected and described, achieving dense object-level understanding and detection. These effects showcase the significant advantages and potential of EagleVision's innovative architecture in the remote sensing vertical field.

\section{Conclusion}
In this paper, we introduce EagleVision, a novel object-level attribute multimodal large language model (MLLM) for remote sensing, seamlessly integrating both object localization and fine-grained attribute comprehension. To enable instruction tuning and performance evaluation of EagleVision, we present the first large-scale remote sensing object attributes understanding dataset, EVAttrs-95K, and the corresponding benchmark, EVBench. Moreover, the Attribute Disentangle module is proposed, ensuring vision token disentanglement for better attribute representation and alignment. Extensive experimental results demonstrate EagleVision achieves state-of-the-art performance in multiple tasks.

{
    \small
    \bibliographystyle{ieeenat_fullname}
    \bibliography{main}
}

\appendix
\clearpage
\maketitlesupplementary

\section{Annotation Process Diagram}
\label{sec:Diagram}
\begin{figure}
  \centering
  \includegraphics[width=0.45\textwidth]{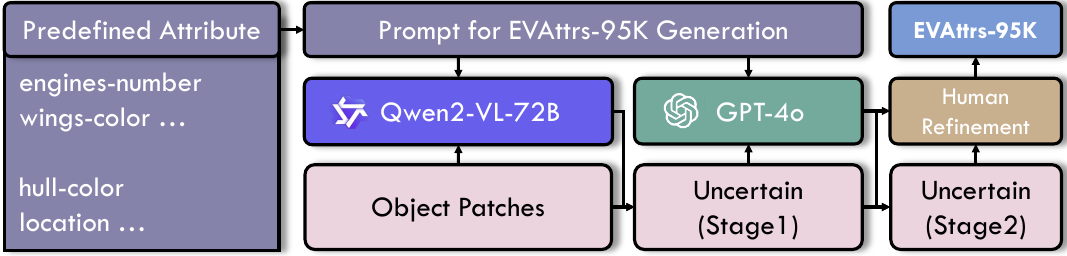}
   \caption{
     \textbf{Annotation process diagram}.
   }
   \label{fig:process}
   \vspace{-3mm}
\end{figure}

The annotation process diagram is shown in Fig. \ref{fig:process}.

\section{Annotation Example}
\label{sec:Example}
\begin{figure}
  \centering
  \includegraphics[width=0.45\textwidth]{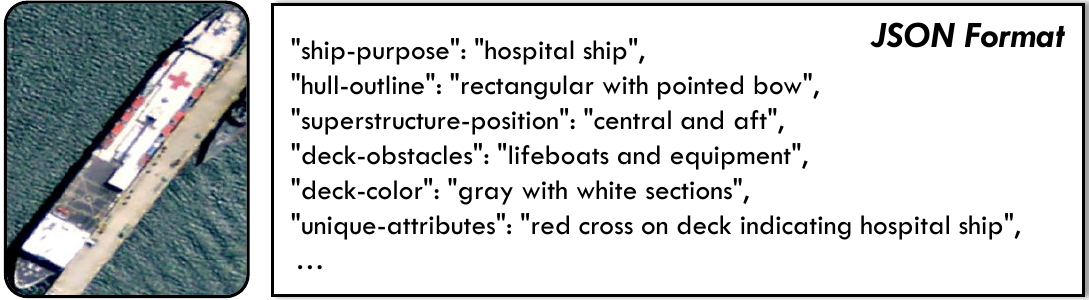}
   \caption{
     \textbf{Annotation example on ShipRSImageNet}.
   }
   \label{fig:example}
   \vspace{-5mm}
\end{figure}

As an example, we provide the attribute annotation result of the ship on ShipRSImageNet, as shown in Fig. \ref{fig:example}.

\section{Predefined Attributes}
\label{sec:Attributes}
The fine-grained attributes of ship and airplane in EVAttrs-95K are shown below. For each existing attribute, we offer an open-end description.

\begin{tcolorbox}[colback=black!5!white,colframe=black!75!black,title=Attributes of Ship,left=3mm,right=1mm,top=2mm,bottom=3mm]
\begin{multicols}{2}
ship-visibility\\ship-purpose\\ship-motion\\ship-capacity\\ship-load-status\\ship-cargo-status\\ship-mooring-status\\hull-color\\hull-size\\hull-shadow\\hull-outline\\superstructure-color\\superstructure-size\\superstructure-height\\superstructure-position\\paint-condition\\bow-design\\stern-design\\deck-utilization\\deck-condition\\deck-obstacles\\deck-color\\deck-structure\\deck-accessories\\passenger-facilities\\container-presence\\container-count\\container-color\\container-layoaut\\container-alignment\\container-densities\\container-type\\machinery-presence\\location\\weather-condition\\water-color\\water-turbulence\\unique-attributes  

\end{multicols}
\end{tcolorbox}

\begin{tcolorbox}[colback=black!5!white,colframe=black!75!black,title=Attributes of Airplane,left=3mm,right=1mm,top=2mm,bottom=3mm]
\begin{multicols}{2}
engine-color\\engine-location\\engine-size\\engine-type\\engines-number\\engines-shape\\engines-visible\\fuselage-color\\fuselage-length\\fuselage-material\\fuselage-shape\\nose-cone-color\\propeller-count\\tail-color\\tail-height\\tail-material\\tail-shape\\tail-type\\wings-angle\\wings-color\\wings-material\\wings-shape\\wings-span\\wings-type
\end{multicols}

\end{tcolorbox}

\section{Prompt Design}
\label{sec:Prompt}
In this paper, we meticulously design three distinct prompts for annotating the EVAttrs-95K dataset, the GPT-assisted evaluation in EVBench and obtaining results from other MLLMs on the OAU task, excluding Eaglevision. The full prompts are provided as follows, with the blue text indicating sections that need to be replaced depending on the objects (ship or airplane).

\begin{tcolorbox}[colback=black!5!white,colframe=black!75!black,title=The Prompt for EVAttrs-95K Generation,top=3mm,bottom=5mm]
Please perform fine-grained visual annotation of the center \textcolor{blue}{ship} in the image based on different attributes (such as body color, body size, etc.).

————————————————————

Please strictly follow these requirements for annotation:

1. All attributes must be consistent with the visual information of the provided image.

2. Provide the confidence level for the annotation, ranging from 0 to 1, where 1 means you are absolutely confident that your annotation is correct.

3. Your annotation description for each attribute must be accurate.

(VERY IMPORTANT) 

4. Your output must conform to JSON format, and only include the following attributes:

confidence

\textcolor{blue}{ship-visibility}

\textcolor{blue}{ship-purpose}

\textcolor{blue}{...}

\end{tcolorbox}

\begin{figure*}
\centering
\begin{tcolorbox}[colback=black!5!white,colframe=black!75!black,title=The Prompt for GPT-assistanted Evaluation (1/2)]

\textbf{System Message}\\[0em]
 
\noindent Please act as an impartial judge and evaluate a multimodal AI assistant's performance in fine-grained attribute understanding. Each data sample includes:
\\[1em]
[Instruction] 

\{Instruction\}\\[0em]

[Reference Answer] 

\{Reference Answer\}\\[0em]

[Assistant's Final Answer] 

\{Assistant's Final Answer\}
\\[1em]
    **Evaluation Criteria:**
    \\[1em]
    1. **Correctness:**\\[0em]
    
    \leftskip=0.4cm
    - **Description:** Ensure that the attribute value in the assistant’s answer is correct and aligns with the reference answer in meaning, allowing for reasonable variations in expression.\\[1em]
    - **Guidelines:**
    
        \leftskip=0.8cm
        - **Accurate:** The attribute value matches the reference answer with minimal errors or reasonable variation in phrasing.
        
        - **Moderate:** The attribute value is misaligned with the reference answer in meaning, or incomplete.
        
        - **Inaccurate:** The attribute value is largely incorrect or misleading, with significant deviation from the reference answer.
        \\[0em]
        
    \leftskip=0.4cm
    - **Notes:**
    
        \leftskip=0.8cm
        - The assistant’s answer should match the reference answer in meaning, even if the wording is different. For example, phrases like "none", "not visible", "invisible", "minimal", and similar expressions can be considered equivalent if they convey the same underlying concept of absence or near-absence.
        
        - If the reference answer implies an absence or a near-absence (e.g., "none", "minimal", "slight"), the assistant’s interpretation should be flexible enough to accommodate slight differences in wording as long as the intended meaning remains clear.
        
        - The assistant can use different terminology to describe the same concept, but if the change in phrasing distorts the original meaning or causes ambiguity, it should be flagged as incorrect.
        
        - Avoid penalizing the assistant for using reasonable variants unless it leads to misunderstanding or over-complication of the reference meaning.
        
        - If an attribute is omitted or missing in the assistant's answer, assess whether the absence can be logically inferred or if it is critical to the answer.\\[0em]
        
    \leftskip=0cm
    2. **Expressiveness:**\\[0em]
    
    \leftskip=0.4cm
    - **Description:** Assess whether each attribute's value sufficiently conveys the necessary information, matching the level of detail required by the reference answer.\\[1em]
    - **Guidelines:**
    
        \leftskip=0.8cm
        - **Adequate:** The value is clear and conveys the required information effectively, whether long, short or different variants.
        
        - **Insufficient:** The value is vague, too brief, or fails to clearly express the necessary information.

\end{tcolorbox}
\end{figure*}
\begin{figure*}
\centering
\begin{tcolorbox}[colback=black!5!white,colframe=black!75!black,title=The Prompt for GPT-assisted Evaluation (2/2)]

    **Scoring Guidelines:**\\[0em]

    - **Scale:** 1 to 5\\[0em]
    
    - **5:** Excellent 
    
        \leftskip=0.4cm
        - The attribute value is highly accurate or with alternative phrasing, match the reference answer in meaning, and clearly convey the required information, even if the phrasing differs slightly.\\[0em]
        
    \leftskip=0cm
    - **4:** Good 
    
        \leftskip=0.4cm 
        - The attribute value is mostly correct, with minor discrepancies in meaning, and still convey the necessary information effectively.\\[0em]
        
    \leftskip=0cm
    - **3:** Satisfactory 
    
        \leftskip=0.4cm 
        - The attribute value is with noticeable errors or unclear expression that weakens the conveyed meaning.\\[0em]
        
    \leftskip=0cm
    - **2:** Needs Improvement 
    
        \leftskip=0.4cm 
        - The attribute value is mostly incorrect, or incomplete, and fail to clearly convey the similar information in reference answer.\\[0em]
        
    \leftskip=0cm
    - **1:** Poor 
    
        \leftskip=0.4cm 
        - The attribute value is completely incorrect or misleading, cannot be understood as a variant of the reference answer, and fails to provide meaningful or necessary information.\\[0em]
        
    \leftskip=0cm
    **Instructions:**\\[0em]

    1. **Assess Correctness, and Expressiveness:** Assess Correctness and Expressiveness for each attribute based on the criteria above.\\[0em]
    
    2. **Attribute Score:** Provide a score for each attribute (1 to 5).\\[0em]
    
    3. **Explanation:** Briefly justify the score for each attribute, especially if there are reasonable variations in phrasing or missing attributes that can be inferred.\\[0em]

    **Output Format:**\\[0em]

    [Explanation] {Your evaluation}

    [attribute\_name 1] \{1-5\}
    
    [attribute\_name 2] \{1-5\}
    
    ...\\[0em]

    \textbf{User}\\[0em]

    [Instruction]
    
    This image includes a remote sensing object in a bird's-eye view. Please help me to explain the visual content of this object in a fine-grained manner.\\[0em]
    
    [Reference Answer]
    
    \textcolor{blue}{\{“ship-visibility”: “Partially Obstructed”, “ship-purpose”: “Military (likely a destroyer based on structure)”, ...\}}\\[0em]
    
    [Assistant's Final Answer]
    
    \textcolor{blue}{\{“ship-visibility”: “partial”, “ship-purpose”: “naval or military operations”, ...\}}

\end{tcolorbox}
\end{figure*}

\begin{figure*}
\centering
\begin{tcolorbox}[colback=black!5!white,colframe=black!75!black,title=The Prompt for OAU Results Generation,bottom=5mm]

    Task Requirements:\\[0em]

  \leftskip=0.2cm 
  Please help me detect all the \textcolor{blue}{ships} in the image, determine their positions, and provide the coordinates of the four corners of their rotated bounding boxes with decimal precision. Identify the class of each \textcolor{blue}{ship} and provide fine-grained attribute descriptions based on the given categories and attributes.\\[0em]

  \leftskip=0cm 
  Output Format:\\[0em]
  
    \leftskip=0.2cm 
    The output must be in the JSON list format.\\[0em]

  \leftskip=0cm 
  \textcolor{blue}{Ship Classes:}\\[0em]
    
    \leftskip=0.2cm 
    List of all possible \textcolor{blue}{ship} classes.
    
    \leftskip=0.2cm 
    Options: \textcolor{blue}{[“Container Ship", “Enterprise", “Container Ship”, “Enterprise”, “Tugboat”, “Cargo”, ...]}\\[0em]

  \leftskip=0cm 
  \textcolor{blue}{Ship Attributes:}\\[0em]
    
    \leftskip=0.2cm
    List of all possible fine-grained \textcolor{blue}{ship} attributes.
    
    \leftskip=0.2cm
    Options: \textcolor{blue}{[“ship-visibility”, “ship-purpose”, “ship-motion”, “ship-capacity”, “ship-load-status”, ...]}\\[0em]

  \leftskip=0cm 
  Output Requirements:\\[0em]
    
      \leftskip=0.2cm
      1. Each \textcolor{blue}{ship} must return its **class**, **position** (with decimal precision), and **attributes** (may contain multiple attributes).\\[0em]
      
      \leftskip=0.2cm
      2. The **class** and **attributes** should be chosen from the provided options.\\[0em]
      
      \leftskip=0.2cm
      3. The **position** must represent the rotated bounding box of the \textcolor{blue}{ship}, with the four corner coordinates [x1, y1, x2, y2, x3, y3, x4, y4] given in decimal precision.\\[0em]
      
      \leftskip=0.2cm
      4. The **attributes** should include the fine-grained descriptions based on the \textcolor{blue}{ship's} visual characteristics.

\end{tcolorbox}
\end{figure*}

For more convenient display, the prompt used for OAU results generation in Fig. \ref{intro} cancels the format restrictions and coordinate output requirements, and supplements the image-level description instruction.

\section{Implementation Details}
\label{sec:parameters}

For EagleVision with different sizes in multiple datasets, we determine their learning rates for training, following the basic principle, the larger model with the lower learning rate. In addition, except FAIR1M adopts a lower language loss weight $\lambda_q$, all other weights are the default 1.0.

\begin{table}
    \centering
    \renewcommand{\arraystretch}{1.0}
    \begin{tabular}{lccccccc}
        \toprule
        
        Dataset          & Size & lr & $\lambda_d$   & $\lambda_o$ & $\lambda_a$ & $\lambda_q$     & n \\
        \midrule
        \multirow{4}{*}{ShipRS}  & 1B & 6e-4 &\multirow{4}{*}{1.0} &\multirow{4}{*}{1.0} &\multirow{4}{*}{1.0} &\multirow{4}{*}{1.0} &\multirow{4}{*}{64} \\\
           & 2B & 4e-4 &  &  &  &  &  \\
           & 4B & 1e-4 &  &  &  &  &  \\
           & 7B & 9e-5 &  &  &  &  &  \\
        \midrule
       \multirow{4}{*}{MAR20}  & 1B & 9e-4 &\multirow{4}{*}{1.0} &\multirow{4}{*}{1.0} &\multirow{4}{*}{1.0} &\multirow{4}{*}{1.0} &\multirow{4}{*}{64} \\\
           & 2B & 6e-4 &  &  &  &  &  \\
           & 4B & 6e-4 &  &  &  &  &  \\
           & 7B & 1e-4 &  &  &  &  &  \\
        \midrule
        \multirow{4}{*}{FAIR1M}  & 1B & 1e-4 &\multirow{4}{*}{1.0} &\multirow{4}{*}{1.0} &\multirow{4}{*}{1.0} &\multirow{4}{*}{0.1} &\multirow{4}{*}{64} \\\
           & 2B & 1e-4 &  &  &  &  &  \\
           & 4B & 6e-5 &  &  &  &  &  \\
           & 7B & 6e-5 &  &  &  &  &  \\
        \bottomrule
    \end{tabular}
    \caption{\textbf{The hyperparmeters for EagleVision.}}
    \vspace{-5mm}
    \label{params}
\end{table}

\section{Additional Visualization Results}
\begin{figure*}
\begin{tcolorbox}[colback=black!5!white, colframe=black!75!black, title=Visualization Results (1/6)]
  \centering
  \includegraphics[width=1.0\textwidth]{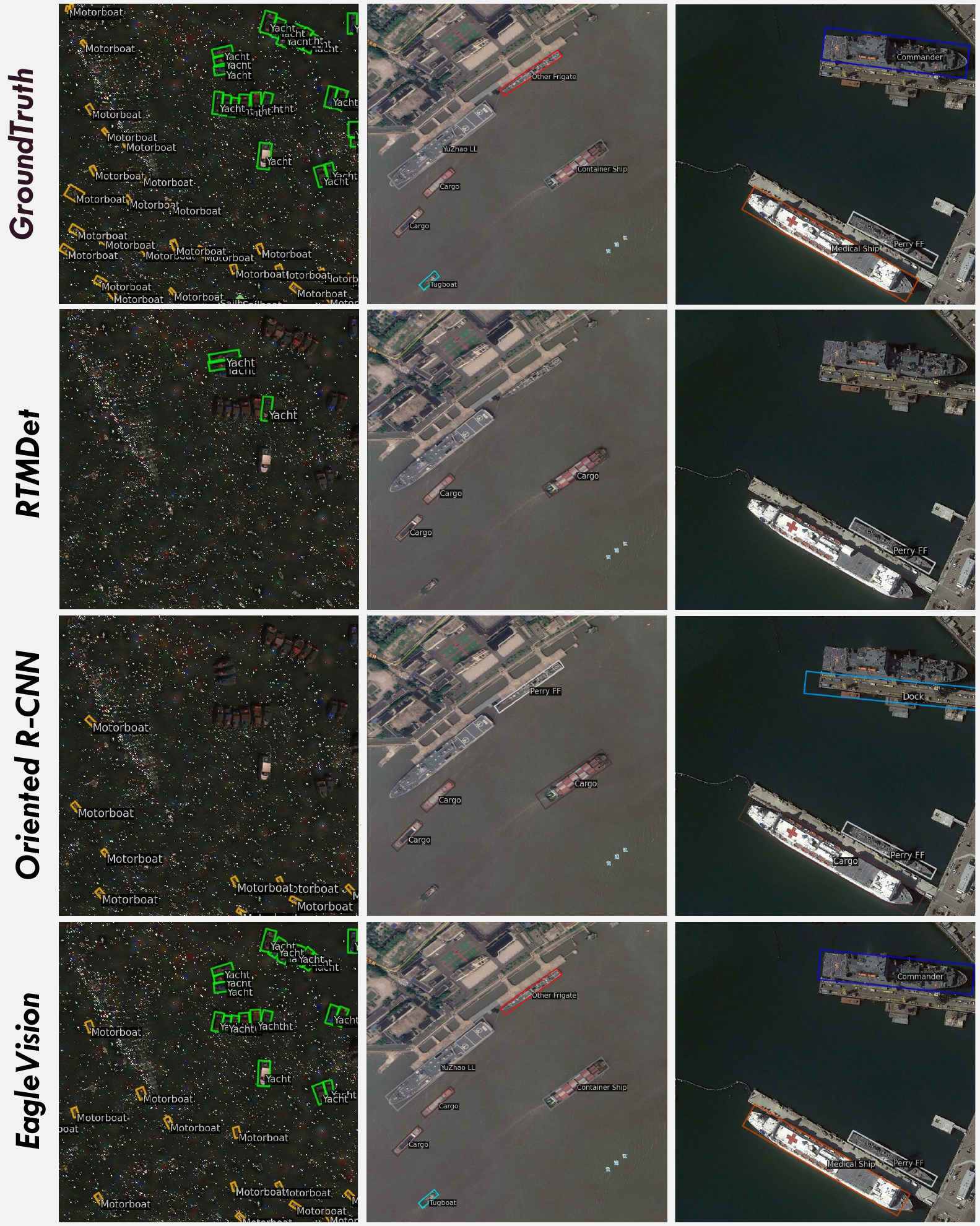}
   \caption{
     \textbf{Visualization results for object detection on ShipRSImageNet dataset}.
   }
   \label{vis1}
\end{tcolorbox}
\end{figure*}

\begin{figure*}
\begin{tcolorbox}[colback=black!5!white, colframe=black!75!black, title=Visualization Results (2/6)]
  \centering
  \includegraphics[width=1.0\textwidth]{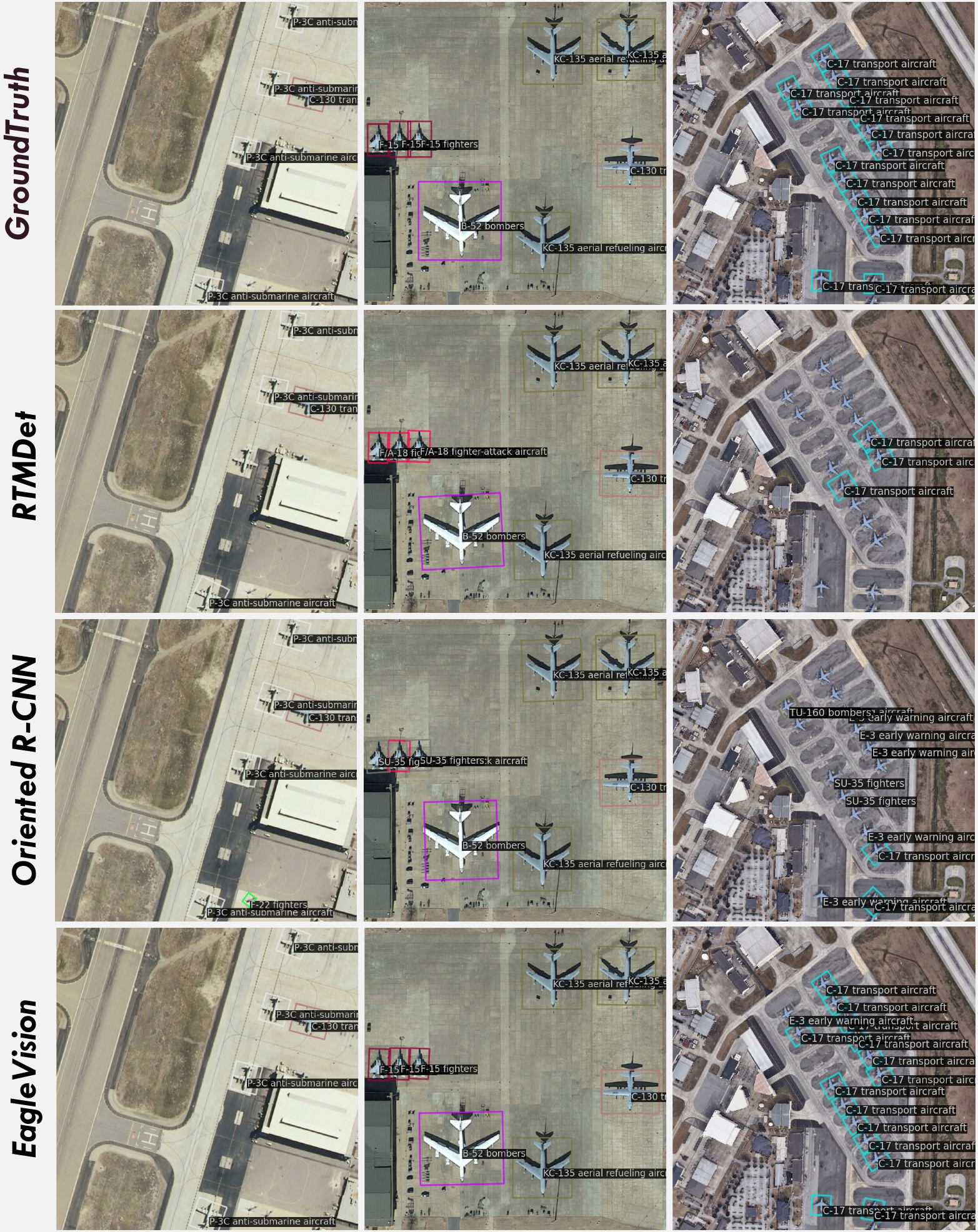}
   \caption{
     \textbf{Visualization results for object detection on MAR20 dataset}.
   }
   \label{vis2}
\end{tcolorbox}
\end{figure*}

\begin{figure*}
\begin{tcolorbox}[colback=black!5!white, colframe=black!75!black, title=Visualization Results (3/6)]
   \centering
   \includegraphics[width=1.0\textwidth]{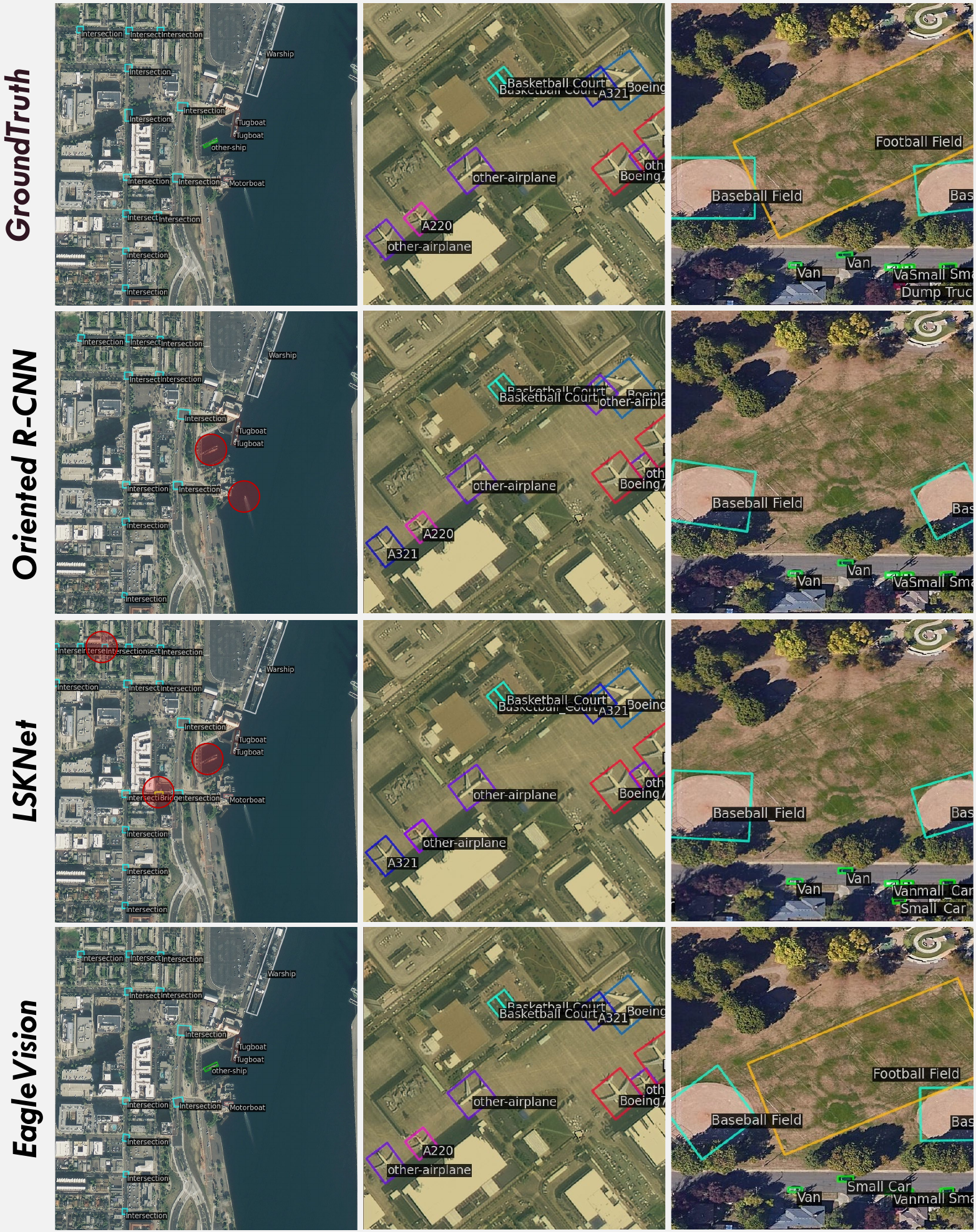}
   \caption{
     \textbf{Visualization results for object detection on the validation set of FAIR1M-v1.0 dataset}. 
   }
   \label{vis3}
\end{tcolorbox}
\end{figure*}

\begin{figure*}
\begin{tcolorbox}[colback=black!5!white, colframe=black!75!black, title=Visualization Results (4/6)]
   \centering
   \includegraphics[width=0.89\textwidth]{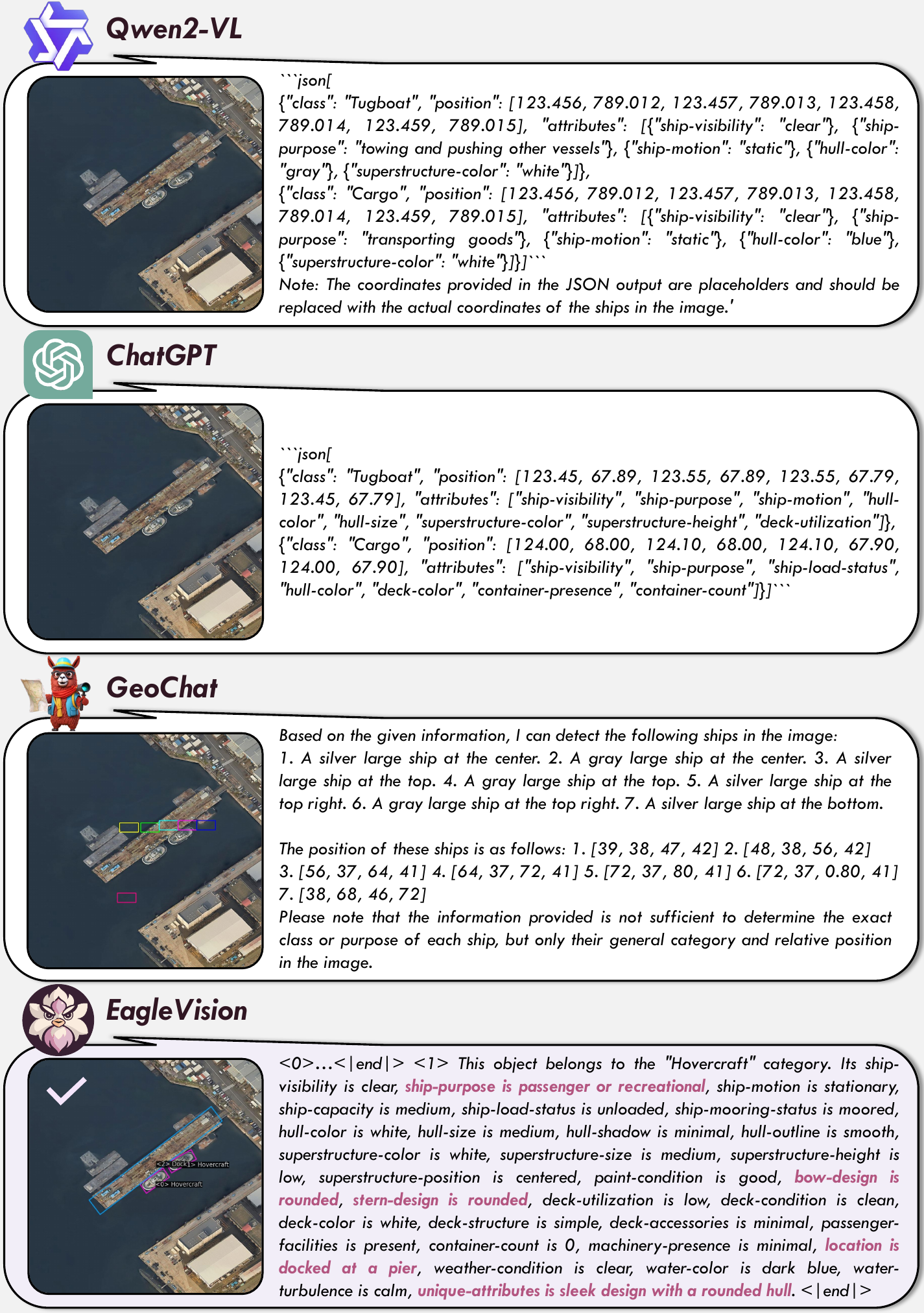}
   \caption{
     \textbf{Visualization results for object attribute understanding on the ShipRSImageNet dataset}. 
   }
   \label{vis4}
\end{tcolorbox}
\end{figure*}

\begin{figure*}
\begin{tcolorbox}[colback=black!5!white, colframe=black!75!black, title=Visualization Results (5/6)]
   \centering
   \includegraphics[width=0.89\textwidth]{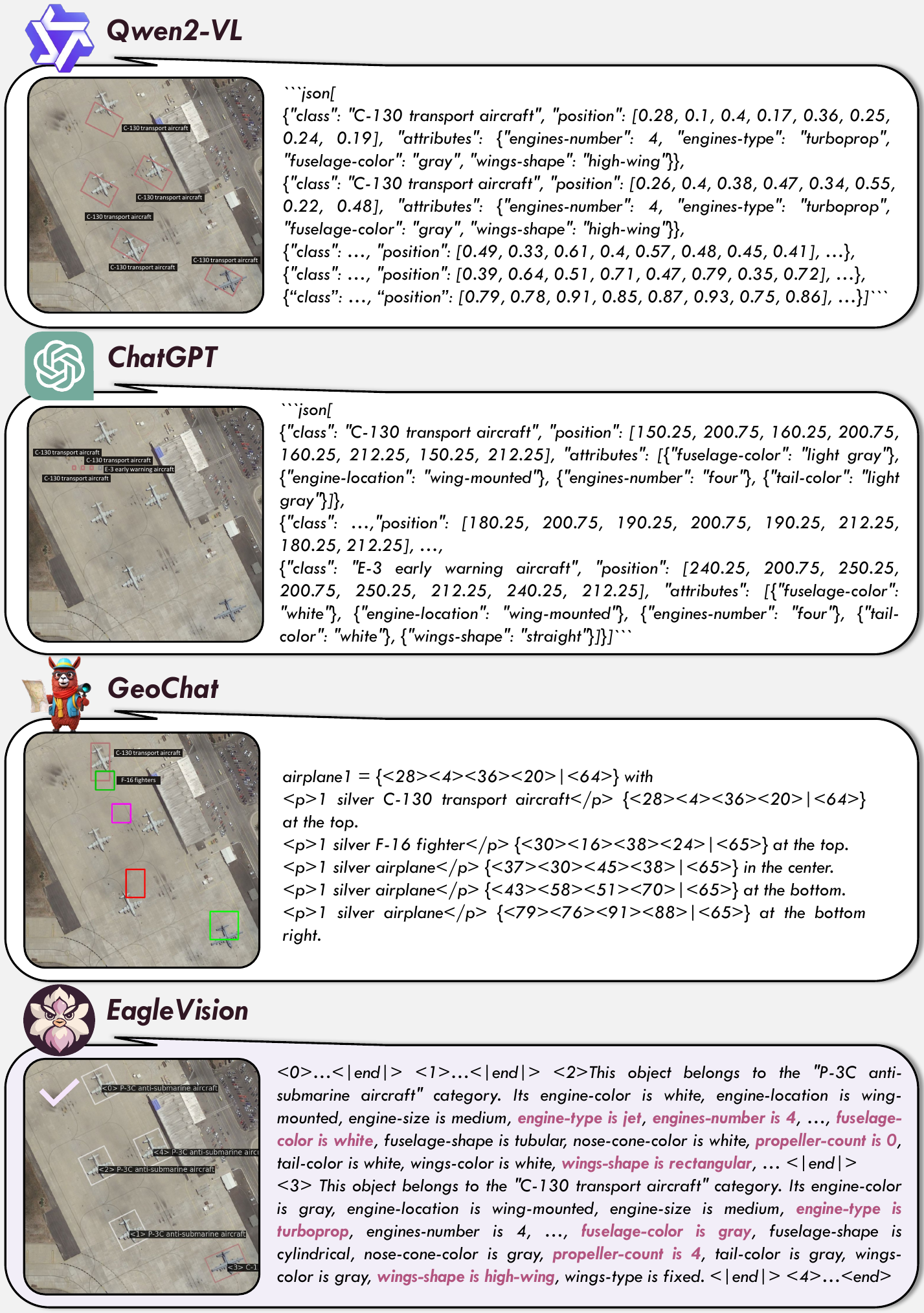}
   \caption{
     \textbf{Visualization results for object attribute understanding on the MAR20 dataset}. 
   }
   \label{vis5}
\end{tcolorbox}
\end{figure*}

\begin{figure*}
\begin{tcolorbox}[colback=black!5!white, colframe=black!75!black, title=Visualization Results (6/6)]
   \centering
   \includegraphics[width=0.89\textwidth]{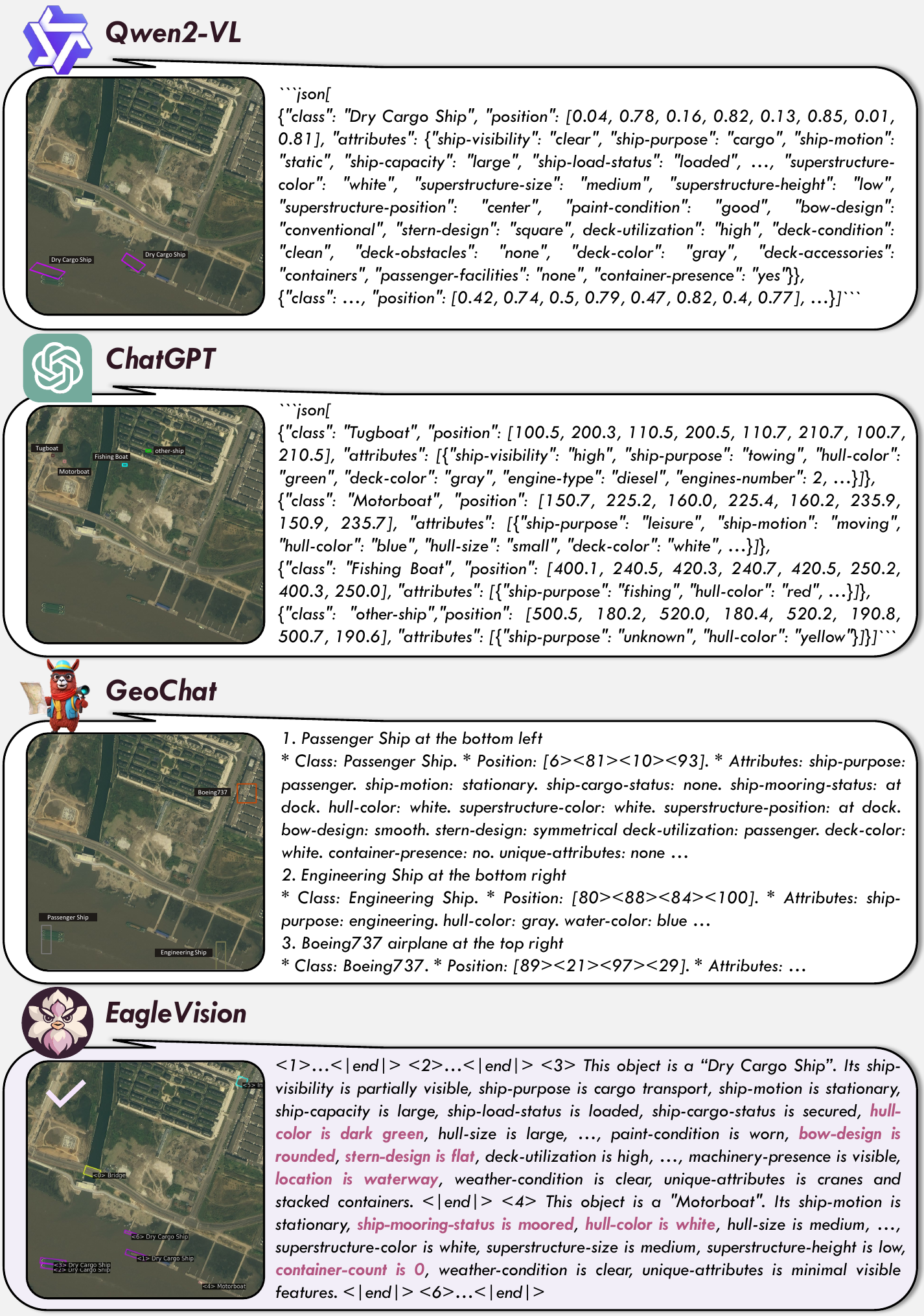}
   \caption{
     \textbf{Visualization results for object attribute understanding on the validation set of FAIR1M-v1.0 dataset}. 
   }
   \label{vis6}
\end{tcolorbox}
\end{figure*}

In this section, we provide more visualization and comparison results for object detection in Fig. \ref{vis1}, \ref{vis2}, and \ref{vis3}, and for object attribute understanding in Fig. \ref{vis4}, \ref{vis5}, and \ref{vis6}.

\end{document}